\documentclass[letterpaper]{article} 
\usepackage{aaai2026}  
\usepackage{times}  
\usepackage{helvet}  
\usepackage{courier}  
\usepackage[hyphens]{url}  
\usepackage{graphicx} 
\usepackage{natbib}  
\usepackage{caption} 
\usepackage{algorithm}
\usepackage{algorithmic}

\usepackage{newfloat}
\usepackage{listings}
\DeclareCaptionStyle{ruled}{labelfont=normalfont,labelsep=colon,strut=off} 
\floatstyle{ruled}
\newfloat{listing}{tb}{lst}{}
\floatname{listing}{Listing}
\title{Introducing Visual Scenes and Reasoning: A More Realistic Benchmark for Spoken Language Understanding}
\author{
    Di Wu\textsuperscript{\rm 1,2}\thanks{~This work was completed during the internship.}, Liting Jiang\textsuperscript{\rm 1}, Ruiyu Fang\textsuperscript{\rm 2}, Bianjing\textsuperscript{\rm 1}, Hongyan Xie\textsuperscript{\rm 3}, Haoxiang Su\textsuperscript{\rm 1}, Hao Huang\textsuperscript{\rm 1,4}\thanks{~Corresponding author.}, Zhongjiang He\textsuperscript{\rm 2}, Shuangyong Song\textsuperscript{\rm 2}, Xuelong Li\textsuperscript{\rm 2}\footnotemark[2] 
}
\affiliations{
    \textsuperscript{\rm 1}School of Computer Science and Technology, Xinjiang University\\
    \textsuperscript{\rm 2}Institute of Artificial Intelligence (TeleAI), China Telecom Corp Ltd\\
    \textsuperscript{\rm 3}School of Computer, Beijing University of Aeronautics and Astronautics\\
    \textsuperscript{\rm 4}Joint International Research Laboratory of Silk Road Multilingual Cognitive Computing

}

\usepackage{bibentry}

\begin{document}

\maketitle

\begin{abstract}
Spoken Language Understanding (SLU) consists of two sub-tasks: intent detection (ID) and slot filling (SF). Given its broad range of real-world applications, enhancing SLU for practical deployment is increasingly critical. Profile-based SLU addresses ambiguous user utterances by incorporating context awareness (CA), user profiles (UP), and knowledge graphs (KG) to support disambiguation, thereby advancing SLU research toward real-world applicability. However, existing SLU datasets still fall short in representing real-world scenarios. Specifically, (1) CA uses one-hot vectors for representation, which is overly idealized, and (2) models typically focuses solely on predicting intents and slot labels, neglecting the reasoning process that could enhance performance and interpretability. To overcome these limitations, we introduce VRSLU, a novel SLU dataset that integrates both \textbf{V}isual images and explicit \textbf{R}easoning. For over-idealized CA, we use GPT-4o and FLUX.1-dev to generate images reflecting users’ environments and statuses, followed by human verification to ensure quality. For reasoning, GPT-4o is employed to generate explanations for predicted labels, which are then refined by human annotators to ensure accuracy and coherence. Additionally, we propose an instructional template, LR-Instruct, which first predicts labels and then generates corresponding reasoning. This two-step approach helps mitigate the influence of reasoning bias on label prediction. Experimental results confirm the effectiveness of incorporating visual information and highlight the promise of explicit reasoning in advancing SLU.
\end{abstract}

\begin{links}
    \link{Datasets}{https://github.com/Tele-AI/TeleVRSLUBench}
\end{links}

\section{Introduction}
Spoken Language Understanding (SLU) is a fundamental component of task-oriented dialogue (TOD) systems \cite{ni2023recent}. It generally comprises two sub-tasks: intent detection (ID) and slot filling (SF) \cite{tur2011spoken}. ID aims to identify the user’s underlying intent, while SF focuses on extracting entities associated with that intent. Together, the predicted intent and slot labels form the basis for generating appropriate system responses and executing subsequent actions within TOD systems. Given its wide applicability in domains such as smart speakers, virtual assistants, and smart home systems, SLU has attracted increasing research attention in recent years \cite{qin2021survey,weld2022survey,muhammad2025joint}.
\begin{figure}[t]
\centering
\includegraphics[width=1.0\columnwidth]{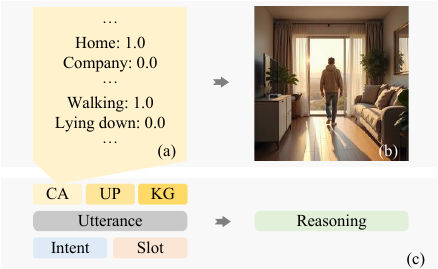}
\caption{The construction process of scene images and reasoning in VRSLU: scene images (b) are generated based on the one-hot vectors (a) in CA, while reasoning are constructed by integrating the utterance, CA, UP, KG, intent, and slot labels (c).}
\label{fig:intro}
\end{figure}

Previous research on SLU can be broadly categorized into two stages. The first stage focuses on traditional SLU under idealized conditions, where user utterances are clear, explicit, and unambiguous. In such settings, models are able to accurately identify intent and slot information based solely on the utterance. Existing methods have already achieved promising results under these conditions \cite{qin2021co,chen2022towards,wu2024cea}.
However, this scenario does not reflect the complexity of real-world applications, where user utterances are often vague or ambiguous. For example, given the utterance ``{\it Play Martial Universe by Tiancan Tudou on the smart screen}", it is difficult for the model to determine whether the user is requesting music, a video, or an audiobook.
To address such challenges, \citet{xu2022text} demonstrated that traditional SLU methods struggle with ambiguous inputs and proposed Profile-based SLU (ProSLU). This task requires the model to integrate additional sources of information to accurately infer user intent from ambiguous utterances. These sources include: Context Awareness (CA), which reflects the user’s current environment and state; User Profile (UP), which captures personal preferences and attributes; and Knowledge Graphs (KG), which provide knowledge about entities in the utterance. 
Although ProSLU marks an important step toward handling utterance ambiguity in real-world scenarios, significant discrepancies still remain between current research efforts and practical deployment:

\begin{itemize}

    \item \textbf{Overly idealized representation of CA}: 
    In ProSLU, CA is represented using one-hot vectors (e.g., ``{\it Home: 1.0, ... Walking: 1.0, ...}"), as illustrated in Figure~\ref{fig:intro}(a). However, such representations are overly simplified and fail to capture the complexity of real-world environments. In practice, CA is more realistically derived from visual inputs, where the model must infer the user’s surroundings and state from images, as shown in Figure~\ref{fig:intro}(b), rather than relying on predefined categorical encodings.

    \item \textbf{Lack of explicit reasoning}: 
    Traditional SLU and ProSLU systems produce only intent and slot predictions without providing the underlying reasoning process. This limitation overlooks the potential benefits of reasoning for improving both accuracy and interpretability. Explicit reasoning not only facilitates a deeper understanding of user intent but also enhances the transparency and trustworthiness of SLU systems, which is crucial for real-world deployment.

\end{itemize}

To address the two challenges discussed above and advance SLU research toward real-world applicability, we construct a new SLU dataset that incorporates both \textbf{V}isual scenes and \textbf{R}easoning, VRSLU.
To mitigate the issue of overly idealized CA representation, we first empirically demonstrate that CA significantly enhance the model’s understanding of user utterances, underscoring their importance. We further argue for the use of images instead of one-hot vectors. Then, we use GPT-4o to convert the one-hot encoded CA vectors into descriptive paragraphs, and generate corresponding images using FLUX.1-dev\footnote{\url{https://huggingface.co/black-forest-labs/FLUX.1-dev}}. All generated images undergo manual verification to ensure semantic alignment and visual quality. 
To address the lack of explicit reasoning, we input the utterance, CA, UP, KG, and the corresponding intent and slot labels into GPT-4o, prompting it to generate explanatory reasoning for each label, as illustrated in Figure~\ref{fig:intro}(c). All reasoning outputs are manually reviewed and refined to ensure both accuracy and clarity.
Furthermore, we propose LR-Instruct, an instructional method that first predicts labels and subsequently generates reasoning. This two-step approach effectively mitigates the impact of reasoning deviations on label prediction in multimodal large language models (MLLMs). Experimental results demonstrate the effectiveness of our approach across multiple MLLMs. Finally, we show that both visual images and explicit reasoning play critical roles in accurately understanding user requests.

The contributions of this work are as follows:

1. To extend SLU to more practical scenarios, we construct a novel dataset, VRSLU, which is the first to include both visual scene images and explicit reasoning processes.

2. We propose LR-Instruct, an instruction-based method that enhances performance across multiple MLLMs on VRSLU. Additionally, we demonstrate the critical role of visual context and the potential of explicit reasoning in accurately identifying intent and slot labels.

\section{Related Work}
\subsection{Traditional SLU Dataset Progress}
The Airline Travel Information Systems (ATIS) dataset has gained significant popularity in the NLP community since its release \cite{hemphill1990atis}, and has been extended to various languages \cite{upadhyay2018almost,xu2020end}. However, due to its single-domain focus on aviation, \citet{tur2011spoken} argues that it is difficult to learn transferable knowledge applicable to other domains from this dataset. Similarly, CAIS, the first dataset primarily targeting the music domain, suffers from comparable limitations \cite{liu2019cm}.
The SNIPS benchmark, collected via crowdsourcing by Snips \cite{coucke2018snips}, contains utterances from voice assistants and smart home devices across seven domains. Its multi-domain nature has contributed to its widespread adoption in SLU research \cite{song2025zero}. Furthermore, \citet{saade2019spoken} expanded the scope of SLU by introducing the Spoken Language Understanding Resource Package (SLURP) \cite{bastianelli2020slurp}, which covers an even broader range of domains.

\subsection{ProSLU Dataset Progress}
\citet{xu2022text} demonstrated that many studies based on the aforementioned datasets fail to effectively handle ambiguous utterances \cite{li2022understanding,zhou2021pin}. To address this issue, they proposed ProSLU, a dataset that incorporates not only user utterances as input to the model, but also integrates UP, CA, and KG information to enhance the model’s ability to accurately interpret user requests \cite{pham2024jpis,teng2024pro}. Compared with traditional SLU datasets, ProSLU is more representative of real-world SLU scenarios.

However, a significant disconnect remains between ProSLU and real-world SLU scenarios, primarily due to overly idealized CA and a lack of explicit reasoning. To address these limitations, we introduce VRSLU, a new dataset designed to overcome the shortcomings of existing SLU benchmarks. Key differences are summarized in Table \ref{table:dataset}.

\begin{table}[h]
\setlength{\tabcolsep}{3.4mm}
\renewcommand\arraystretch{1.1}
\centering
\begin{tabular}{l|ccc}
\hline
\hline
Model   &  Ambiguous? & Visual? & Reasoning? \\
\hline
ATIS & $\times$ & $\times$ & $\times$ \\
CAIS & $\times$ & $\times$ & $\times$\\
SNIPS & $\times$ & $\times$ & $\times$ \\
SLURP & $\times$ & $\times$ & $\times$ \\
ProSLU & $\surd$ & $\times$ & $\times$ \\
VRSLU & $\surd$ & $\surd$ & $\surd$ \\
\hline
\hline
\end{tabular}
\caption{Differences between the VRSLU and previous SLU datasets. Here, $\times$ denotes the absence of a particular characteristic in the dataset, while $\surd$ indicates its presence.}
\label{table:dataset}
\end{table}

\subsection{Advances in LLMs for SLU} 
LLMs have demonstrated potential across a wide range of NLP tasks \cite{he2024telechat,li2024tele,wang2024telechat,li202452b,wang2025technical}, including SLU. Previous SLU approaches \cite{wu2024dual,aimaiti2024uyghur} typically formulate ID as a text classification task \cite{li2022survey,xiong2024dual}, and SF as a sequence labeling task \cite{qu2023survey}. In contrast, most LLM-based methods treat SLU as a generative task. 
\citet{pan2023preliminary} explores the zero-shot performance of ChatGPT on SLU. Subsequently, \citet{zhu2024zero} proposes GPT-SLU, which models the relationship between ID and SF through prompting.
Similarly, \citet{teng2024pro} applied GPT to the ProSLU dataset. 

Due to the lack of reasoning annotations, existing models generate only intent and slot labels, limiting the generative potential of LLMs. Moreover, applying MLLMs to SLU remains underexplored because few SLU datasets include visual information.

\section{Visual Scene Construction}
\subsection{Significance of CA}
\begin{figure}[h]
\centering
\includegraphics[]{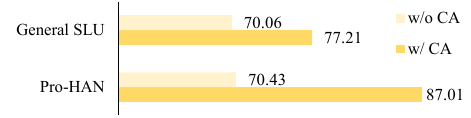}
\caption{Overall accuracy of General SLU and ProHAN under {\it w/ CA} and {\it w/o CA} Settings.}
\label{fig:ca_pre}
\end{figure}

Users’ requests are often influenced by their surrounding environment and current state. To further validate the importance of CA, we conduct experiments on the ProSLU dataset using both General SLU \cite{xu2022text} and ProHAN \cite{teng2024pro} under two settings: with CA information ({\it w/ CA}) and without CA information ({\it w/o CA}). The results are presented in Figure~\ref{fig:ca_pre}. 

As shown in the figure, both models exhibit a decrease in overall accuracy under the {\it w/o CA} setting compared to the {\it w/ CA} setting. This indicates that incorporating CA information enables the model to better complete ID and SF.

\subsection{Necessity of Visual Representation}
As shown in Figure~\ref{fig:intro}(a), CA in ProSLU is represented as a one-hot vector. However, in real-world scenarios, conveying the user’s surrounding environment and state in this manner is impractical. A more realistic setting involves providing the model with an image that captures this context, from which it can infer the relevant CA information. Learning CA from images helps narrow the divide between research and practical deployment.

\subsection{Visual Scene Construction Process}
Due to the high costs associated with capturing real user scenarios and states, which involve diverse environments, camera angles, and recruiting users of various ages and genders, we use FLUX.1-dev to directly generate images without specifying scene configurations, camera perspectives, or user attributes. This approach enables us to obtain a wide variety of scene images while keeping costs manageable.

\begin{figure}[h]
\centering
\includegraphics[]{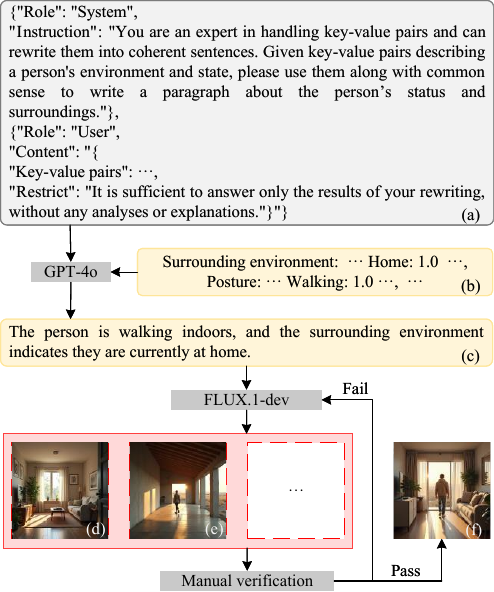}
\caption{The process of image construction. Where, (a) shows the prompt, (b) represents CA one-hot vectors, (c) is the corresponding descriptive paragraph, (d) and (e) are invalid images, highlighted in red, and (f) is a valid image.}
\label{fig:v_process}
\end{figure}

Initially, we attempt to generate images directly from CA one-hot vectors using FLUX.1-dev; however, the results were unsatisfactory. Consequently, we adopt an alternative approach: first, GPT-4o generates descriptive paragraphs based on the CA information, and then FLUX.1-dev produces images from these descriptions. The overall process is illustrated in Figure~\ref{fig:v_process}. (Note: The GPT-4o version used in this work is dated 2023-05-15, with a temperature setting of 0.7. The FLUX.1-dev hyperparameters are as follows: guidance scale is set to 3.5 and the number of inference steps is 50. The scene images are generated on a single NVIDIA A6000 GPU with 48GB of memory.)

\begin{figure*}[h]
\centering
\includegraphics[]{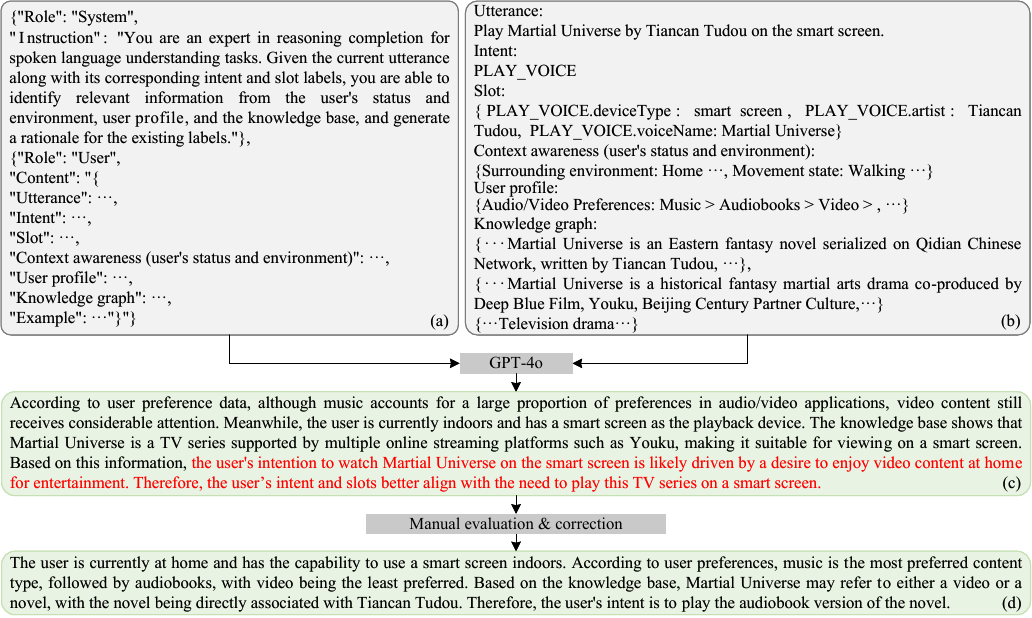}
\caption{The process of reasoning construction. Where, (a) is the prompt, (b) is an example (information irrelevant to the user's request has been omitted for clarity), (c) is the original reasoning generated by GPT-4o with errors highlighted in red, and (d) is the manually corrected result.}
\label{fig:r_process}
\end{figure*}

Due to the high cost of using GPT-4o, we initially sample 100 instances from the dataset to iteratively refine the prompt. We start by designing the initial prompt and using the API to generate descriptive paragraphs for these instances. Subsequently, the paragraphs are used to generate corresponding images with FLUX.1-dev.

For scene image verification, we recruit two professional annotators with prior experience in image annotation to assess the quality of the 100 generated images. The annotators are asked to identify errors such as logical inconsistencies or failures to accurately reflect the user’s environment and state. After multiple rounds of prompt refinement, the final prompt used for image generation is presented in Figure~\ref{fig:v_process}(a). Images generated with this prompt achieve accuracies of 85.00\% and 87.00\% according to the two annotators, respectively. The inter-annotator agreement, measured by Gwet’s AC1, is 0.9170, indicating high reliability.

After finalizing the prompt for GPT-4o, we follow the workflow illustrated in Figure \ref{fig:v_process} to generate a corresponding visual scene for each CA entry and conduct manual verification on all generated images. An image is accepted only if both annotators agree that it meets the predefined quality criteria; otherwise, it is regenerated.
As shown in Figure~\ref{fig:v_process}(d), there is no person in the image, although the scene is a home. In Figure~\ref{fig:v_process}(e), the character is depicted as being neither at home nor in a walking state. Both images fail to accurately reflect CA provided in Figure~\ref{fig:v_process}(b), they are considered invalid.
Invalid images are regenerated using FLUX.1-dev until a valid version is produced.
As illustrated in Figure~\ref{fig:v_process}(f), a valid image is semantically consistent with CA and visually reasonable, thereby satisfying the annotation criteria.

\section{Reasoning Construction}
\subsection{Potential Benefits of Reasoning}
Reasoning is the process of thinking in a logical and systematic manner, leveraging evidence and prior experience to reach conclusions or make decisions \cite{valmeekam2022large}. It enables models to provide valid justifications for their predictions, thereby improving interpretability and transparency \cite{huang2023towards}. Moreover, incorporating appropriate reasoning processes has been shown to enhance model performance across a range of tasks \cite{zhang2023multimodal, wang2024t}.
However, existing SLU research has largely focused on intent and slot label prediction, while neglecting the generation of reasoning processes. This oversight limits both the accuracy and interpretability of models, thereby hindering the deployment of SLU systems in real-world scenarios. A major contributing factor is the lack of annotated reasoning processes in current SLU datasets, which deprives models of clear learning targets.
To address this issue, we construct reasoning explanations for each sample using the following approach.

\subsection{Reasoning Construction Process}
The overall reasoning construction process is illustrated in Figure \ref{fig:r_process}. Specifically, for each sample, we first use GPT-4o to generate an explanation of the user's request, which is then refined through manual evaluation and correction.

To optimize the prompting strategy, we randomly select 100 instances and iteratively refine the prompt used for reasoning generation within our computational constraints. After several rounds of refinement, we determine that the optimal input should include the utterance, intent label, slot labels, UP, KG, and CA (in textual form). 
We exclude previously generated images, as they tend to reduce the quality of the reasoning, likely due to GPT-4o needing to interpret the image in order to extract CA. In contrast, providing CA directly in text form allows for more accurate reasoning. To further enhance quality, we manually construct a reasoning example for a randomly selected instance and include it in the prompt. The final version of the prompt is shown in Figure \ref{fig:r_process}(a).

Although we iteratively refine the prompt based on 100 sampled instances, the generated reasoning is not always coherent. As shown in Figure \ref{fig:r_process}(b), the intent label of the user utterance is ``{\it PLAY\_VOICE}'', and the corresponding slot labels also indicate a voice-related request. While the KG provides relevant information about the novel, GPT-4o still misinterprets the user's intent as playing a video, as illustrated in Figure \ref{fig:r_process}(c). This may be due to the ambiguity of the utterance itself, as well as the presence of irrelevant information in UP, KG, and CA. Furthermore, GPT-4o occasionally hallucinates and fails to correctly interpret the intent and slot labels, resulting in incorrect reasoning in the output.

To address these issues, we invite two annotators with experience in SLU-related labeling tasks (different from those who annotated the images) and two doctoral students specializing in TOD research to manually evaluate and revise the generated reasoning. The procedure is as follows:

1. \textbf{Define Annotation Guidelines}: 
Annotators compare the reasoning generated by GPT-4o with the intent and slot to assess its clarity, relevance, and accuracy. If revision is needed, they must provide justifications, such as vague explanations or weak alignment with the labels. This feedback then informs the discussion and refinement process in 5.

2. \textbf{Group Assignment}: 
Groups are formed by drawing lots. Each group consists of one annotator with experience in SLU and one doctoral student with a background in TOD system research.

3. \textbf{First-round Review}: 
Each group independently reviews and revises the entire dataset. Within each group, the two members collaborate to evaluate and edit the reasoning.

4. \textbf{Cross Review}: 
After the initial review, the two groups exchange their annotated samples. Each group then jointly evaluates the annotations made by the other group.

5. \textbf{Result Integration}: 
Each sample produces two revised versions of the reasoning. If the versions are consistent, they are merged. In case of discrepancies, all four participants engage in a joint discussion to finalize the reasoning.

These steps complete the construction of the reasoning component in the dataset. A revised reasoning example is shown in Figure \ref{fig:r_process}(d). Compared to the original reasoning generated by GPT-4o, the revised version emphasizes that the user's current environment is suitable for using a smart screen, removes the video-related content, clarifies that ``{\it Martial Universe}'' refers to the novel by the author ``{\it Tiancan Tudou}'', and concludes that the user intends to play the audiobook version of the novel.

\begin{table*}[h] 
\setlength{\tabcolsep}{1.9mm}
\renewcommand\arraystretch{1.1}
\centering
\begin{tabular}{l|cccccc}
\hline
\hline
Model   & Intent (Acc) & Slot (F1) & Overall (Acc) & Reasoning (Bleu-1)  & Reasoning (Bleu-4) & Reasoning (Sim) \\
\hline
Qwen2-VL-2B  & 74.01 & 74.42 & 69.11 & 44.93 & 12.57 & 74.79 \\
~~~~~~ + Ours& 78.96 & 76.34 & 72.76 & 45.19 & 12.89 & 76.33 \\
\hline
Qwen2.5-VL-3B & 84.37 & 85.10 & 79.66 & 46.97 &	13.90 & 77.43 \\
~~~~~~ + Ours& 87.19 & 87.37 & 81.54 & 47.21 &	14.02 & 77.56 \\
\hline
Yi-VL-6B  & 80.98 & 82.14 & 77.78 & 47.20 & 14.15 & 77.81\\
~~~~~~ + Ours& 86.93 & 87.14 & 82.86 & 47.66 & 14.73 & 78.66\\
\hline
Qwen2-VL-7B  & 84.56 & 86.60 & 81.54 & 47.17 & 14.56 & 78.12\\
~~~~~~ + Ours& 91.64 & 92.12 & 85.08 & 48.01 & 14.89 & 78.45\\
\hline
Qwen2.5-VL-7B  & 86.44 & 87.91 & 82.86 & 47.91 & 14.72 & 78.76 \\
~~~~~~ + Ours& 92.66 & 92.62 & 86.63 & 48.61 & 14.98 & 78.94 \\
\hline
\hline
\end{tabular}
\caption{Main experiment results.}
\label{table:mian}
\end{table*}

\section{Method} 
\subsection{Task Description}
Following previous research on LLM-based SLU \cite{wu2022incorporating,xing2024dc}, ID and SF are typically formulated as question-answering tasks. Notably, in VRSLU, the model input includes not only the user utterance $U$, but also three types of knowledge: UP $T_{up}$ and KG $T_{kg}$ in textual form, and CA $I_{ca}$ as visual scenes. The model generates the reasoning process $R$ that underlies the user's request, as well as the intent label $L_{I}$ and slot labels $L_{S}$.
\begin{equation}
L_{I}, L_{S}, R = \mathrm{MLLM}(U,T_{up},T_{kg},I_{ca}).
\end{equation}

\subsection{Instructional Template}
VRSLU requires models to predict both intent and slot labels, as well as to generate the corresponding reasoning process. However, deviations in the reasoning process of LLMs can lead to incorrect label predictions \cite{chua2024bias}. To mitigate this, inspired by \citet{hase2021can,lyu2024towards}, we propose LR-Instruct, an instructional template framework that first performs label prediction followed by reasoning generation.
\begin{figure}[h]
\centering
\includegraphics[]{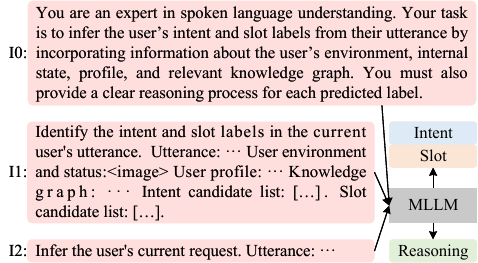}
\caption{
The proposed LR-Instruct for VRSLU. Due to space limitations, we omit the case details as well as the candidate lists for intent and slot labels.}
\label{fig:prompt}
\end{figure}

As illustrated in Figure \ref{fig:prompt}, we construct a multi-turn question-answering prompt consisting of three parts. The first part (I0) instructs the model to role-play as an SLU expert and introduces the task. The remaining two parts correspond to the model’s expected outputs. In I1, the model is required to select appropriate intent and slot labels from the candidate list based on the utterance, the provided CA image, UP, and KG. In I2, the model is asked to provide the reasoning behind the label selection. All three components (I0–I2) are used during both training and inference.

\section{Experiments}
\subsection{Multimodal Large Language Model}
Since VRSLU includes images, we adopt the following MLLMs as baselines in our experiments:
\textbf{Yi-VL-6B} \cite{ai2024yi}. The model employs a Vision Transformer for image encoding, while the projection module facilitates the alignment between image and text.
\textbf{Qwen2-VL} \cite{wang2024qwen2}. It is a multimodal extension of Qwen2-7B that can process visual and language inputs simultaneously. We use two parameter scales for this model: 2B and 7B.
\textbf{Qwen2.5-VL} \cite{bai2025qwen2}. It is an upgraded version of Qwen2-VL.  We use two parameter scales for this mode: 3B and 7B. 
These MLLMs are required to perform reasoning before predicting the intent and slot labels.

\subsection{Experiment Setup}
For fairness, both the proposed LR-Instruct and all baselines are fine-tuned using LoRA \cite{hu2022lora}. We use the AdamW optimizer \cite{loshchilov2017decoupled} with a learning rate of 5e-5 and train each model for 3 epochs, with a per-device batch size of 1. The LoRA rank is set to 8 and the dropout rate to 0.1. To ensure reproducibility, we use a fixed random seed of 42 across all MLLMs. All experiments are conducted on eight NVIDIA A100 GPUs (40GB each).

\subsection{Evaluation Metrics}
Consistent with prior SLU work \cite{wu2025int}, F1-score (F1) is used to evaluate the performance of SF, accuracy (Acc) to evaluate ID performance, and overall accuracy to evaluate sentence-level semantic frame parsing performance. Since VRSLU requires reasoning in the output, we use BLEU-1 and BLEU-4 scores to evaluate reasoning quality. We also compute the cosine similarity between the generated and reference reasoning texts using Qwen3-Embedding-0.6B \cite{zhang2025qwen3}, following the official Sentence Transformers guidelines. 

\subsection{Main Result}
The experimental results of various MLLMs combined with the proposed LR-Instruct are presented in Table \ref{table:mian}. 

1. When combined with various MLLMs, the proposed LR-Instruct consistently improves intent Acc, slot F1, and overall Acc performance. It also achieves notable gains in reasoning metrics, including BLEU-1, BLEU-4, and cosine similarity. A key factor contributing to these improvements is that LR-Instruct requires MLLMs to predict labels prior to generating reasoning, thereby mitigating the negative effects of reasoning errors. Moreover, the improvements achieved by our method are robust across variations in MLLM parameters and architectures, demonstrating its consistent effectiveness.

2. The quality of reasoning is positively correlated with intent and slot prediction accuracy when using the same underlying MLLM. Higher reasoning performance reflects a deeper understanding of the user’s request, which in turn enhances the model’s ability to accurately identify intent and slot labels.

\begin{figure*}[h]
\centering
\includegraphics[]{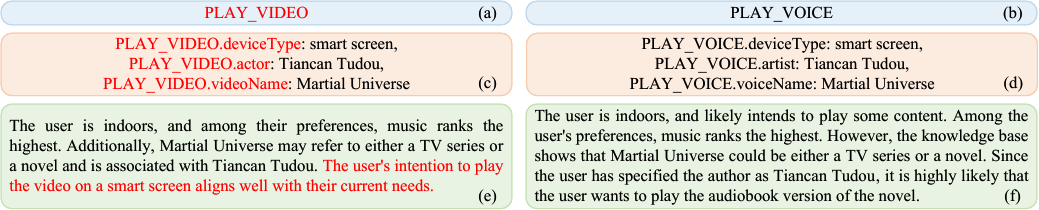}
\caption{Predictions and reasoning from Qwen2.5-VL-7B and Qwen2.5-VL-7B+LR-Instruct. (a), (c), and (e) display the intent, slot, and reasoning of Qwen2.5-VL-7B; (b), (d), and (f) show the corresponding outputs from Qwen2.5-VL-7B+LR-Instruct.}
\label{fig:badcase}
\end{figure*}

\subsection{Analysis}
We construct VRSLU, the first SLU dataset that integrates both visual information and reasoning processes. This raises three key questions:
Q1: Can the constructed scene images effectively reflect the user’s current environment and state, thereby enhancing the model’s understanding of the user’s utterances?
Q2: What are the potential benefits of requiring the model to output not only intent and slot labels but also the corresponding reasoning process?
Q3: What risks might arise from explicitly generating the reasoning process as part of the model’s output?

To investigate the above questions, we carry out the following analyses.

A1: 
As shown in Figure \ref{fig:ca_pre}, the CA one-hot vector helps the model understand user requests. If constructed images are truly effective, removing them or replacing them with low-quality alternatives should degrade model performance. To verify this, Table \ref{table:ablation} presents the performance of our LR-Instruct method (Qwen2.5-VL-7B) under two conditions: without the CA image ({\it w/o $I_{ca}$}) and with CA images that have not been manually verified ({\it w/ unverified $I_{ca}$}). Compared to the original setting, overall accuracy drops by 3.39\% and 1.18\%, respectively. These results suggest that \textbf{manually verified images more accurately represent the user's environment and state, thereby confirming the effectiveness of constructed scene images in enhancing the model’s understanding of user utterances.}

\begin{table}[h]
\setlength{\tabcolsep}{0.6mm}
\renewcommand\arraystretch{1.1}
\centering
\begin{tabular}{l|ccc}
\hline
\hline
Setting  & Intent (Acc) & Slot (F1) & Overall (Acc)\\
\hline
Ours & 92.66 & 92.62 & 86.63  \\
\hline
w/o $I_{ca}$ & 89.08 & 89.78 & 83.24 \\
w/ unverified $I_{ca}$ & 90.45 & 91.24 & 85.45 \\
\hline
w/o R & 90.40 & 91.04 & 85.15 \\
w/ uncorrected R & 89.45 & 90.01 & 83.86 \\
\hline
w/o $I_{ca}$\&R & 87.76 & 87.81 & 80.98 \\
w/ automated $I_{ca}$\&R & 89.45 & 89.24 & 83.05 \\
\hline
\hline
\end{tabular}
\caption{Impact of CA image and reasoning on results. }
\label{table:ablation}
\end{table}

A2: Requiring the model to generate a reasoning process offers two main advantages:

(1) Intuitively, \textbf{requiring the model to generate a reasoning process enhances interpretability}. As shown in Figure~\ref{fig:badcase}, we compare the outputs of Qwen2.5-VL-7B and Qwen2.5-VL-7B+LR-Instruct for the case in Figure~\ref{fig:r_process}(b). While the former correctly identifies ``{\it Tiancan Tudou}'' as a novel or TV series, it ultimately misinterprets the user’s intent as requesting a video. In contrast, the proposed LR-Instruct retrieves the author ``{\it Tiancan Tudou}'' from the knowledge base and grounds this information in the user’s utterance. This enables the model to infer that the user intends to play the audiobook version of the novel, thereby accurately predicting the intent and slot labels.
Regardless of correctness, generating an explicit reasoning process could improve model transparency and facilitate interpretability.

(2) Moreover, \textbf{requiring the model to generate reasoning helps unlock its potential in ID and SF, thereby improving overall performance.} We evaluate two settings: removing the reasoning process ({\it w/o R}) and using uncorrected automatically generated reasoning ({\it w/ uncorrected R}). As shown in Table \ref{table:ablation}, {\it w/o R} leads to drops in intent Acc, slot F1, and overall Acc compared to the vanilla setting. This indicates that requiring reasoning helps the model better understand the utterance and make more accurate predictions. The performance decline in {\it w/ uncorrected R} also suggests that GPT-4o’s generated reasoning still benefits from human refinement.
Additionally, we conduct experiments by removing both CA images and reasoning ({\it w/o $I_{ca}$\&R}), as well as using unverified CA images and uncorrected reasoning ({\it w/ automated $I_{ca}$\&R}). The results show consistent drops across all metrics compared to the vanilla setting, confirming the usefulness of both CA (image) and reasoning, and highlighting the importance of human verification.

\begin{figure}[h]
\centering
\includegraphics[]{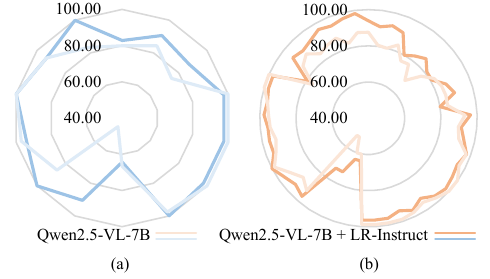}
\caption{
Comparison of results for different categories of intent (a) and slot (b).}
\label{fig:6class}
\end{figure}

A3: As illustrated in Figure~\ref{fig:badcase}, \textbf{a key risk of explicitly generating the reasoning process is that flawed reasoning can negatively impact the accuracy of label predictions}. Supporting this observation, Tables~\ref{table:mian} and \ref{table:ablation} show that Qwen2.5-VL-7B, which generates reasoning before making predictions, performs worse than its counterpart without reasoning ({\it w/o R}). This suggests that if not well controlled, reasoning generation may hinder model performance. The proposed LR-Instruct mitigates this issue by first predicting labels and then generating reasoning, which enhances the model’s ability to discriminate between intent and slot categories. As shown in Figure \ref{fig:6class}, after applying LR-Instruct to the Qwen2.5-VL-7B, its recognition performance on most intent and slot categories surpasses that of the baseline, intuitively demonstrating the effectiveness of our method.

\section{Conclusion}
In this work, we propose a novel benchmark, VRSLU, which is the first SLU dataset that incorporates user environment and statue scene images as part of the input, and requires models to generate reasoning processes. We also propose LR-Instruct, a method that first prompts the model to predict labels before generating the reasoning process, thereby effectively mitigating the influence of reasoning errors on prediction outcomes. Finally, we highlight the contributions of scene images and the potential benefits of explicit reasoning. In future work, we will explore extending VRSLU to AI networks \cite{shao2025ai} to make SLU research more closely aligned with real-world applications. 


\section{Acknowledgment}
The paper was supported by the following projects: National Natural Science Foundation of China (62466055, 61663044), Research Project of the State Language Commission (ZDI145-96), and the Excellent Doctoral Student Research Innovation Project of Xinjiang University (No. XJU2022BS077).

\bibliography{aaai2026}

\end{document}